\documentclass[letterpaper]{article} 
\usepackage{aaai25}  
\usepackage{times}  
\usepackage{helvet}  
\usepackage{courier}  
\usepackage[hyphens]{url}  
\usepackage{graphicx} 
\usepackage{natbib}  
\usepackage{caption} 
\usepackage{algorithm}
\usepackage{algorithmic}

\usepackage{multirow}
\usepackage{amsmath}
\usepackage{newfloat}
\usepackage{listings}
\DeclareCaptionStyle{ruled}{labelfont=normalfont,labelsep=colon,strut=off} 
\floatstyle{ruled}
\newfloat{listing}{tb}{lst}{}
\floatname{listing}{Listing}
\title{RepFace: Refining Closed-Set Noise with Progressive Label Correction for Face Recognition}

\author{
    Jie Zhang\textsuperscript{\rm 1, \rm 2, \rm 3}, Xun Gong\textsuperscript{\rm 1, \rm 2, \rm 3}\thanks{Corresponding author}, Zhonglin Sun\textsuperscript{\rm 4, \rm 1}
}
\affiliations{
    \small
   \textsuperscript{\rm 1} School of Computing and Artificial Intelligence, Southwest Jiaotong University, Chengdu 611756, China\\
   
   \textsuperscript{\rm 2} Engineering Research Center of Sustainable Urban Intelligent Transportation, Ministry of Education, Chengdu 611756, China\\
   
   \textsuperscript{\rm 3} Manufacturing Industry Chains Collaboration and Information Support Technology Key Laboratory of Sichuan Province, Southwest Jiaotong University, Chengdu 611756, China\\
   
   \textsuperscript{\rm 4}Queen Mary university of London, UK\\

}

\usepackage{bibentry}

\begin{document}

\maketitle

\begin{abstract}
Face recognition has made remarkable strides, driven by the expanding scale of datasets, advancements in various backbone and discriminative losses. However, face recognition performance is heavily affected by the label noise, especially closed-set noise. While numerous studies have focused on handling label noise, addressing closed-set noise still poses challenges. This paper identifies this challenge as training isn't robust to noise at the early-stage training, and necessitating an appropriate learning strategy for samples with low confidence, which are often misclassified as closed-set noise in later training phases. To address these issues, we propose a new framework to stabilize the training at early stages and split the samples into clean, ambiguous and noisy groups which are devised with separate training strategies. Initially, we employ generated auxiliary closed-set noisy samples to enable the model to identify noisy data at the early stages of training. Subsequently, we introduce how samples are split into clean, ambiguous and noisy groups by their similarity to the positive and nearest negative centers. Then we perform label fusion for ambiguous samples by incorporating accumulated model predictions. Finally, we apply label smoothing within the closed set, adjusting the label to a point between the nearest negative class and the initially assigned label. Extensive experiments validate the effectiveness of our method on mainstream face datasets, achieving state-of-the-art results. The code will be released upon acceptance. 

\end{abstract}

\section{Introduction}
Face recognition is a prominent research area in pattern recognition, drawing significant attention and advancements in recent years. The application of FR consists of security and surveillance, and mobile unlocking~\textit{et al}. The remarkable progress in face recognition technology can be largely attributed to three key factors: the emergence of vast, large-scale face datasets~\cite{yi2014learning, guo2016ms, kemelmacher2016megaface, cao2018vggface2, zhu2022webface260m, liu2015faceattributes}, the pivotal role played by backbones~\cite{he2016deep, howard2019searching, tan2019efficientnet, dosovitskiy2020image,sun2022part,sun2024lafs}, and the advancement in loss functions~\cite{liu2017sphereface, deng2019arcface, wang2018cosface, kim2022adaface, wang2020mis, huang2020curricularface, wen2021sphereface2, saadabadi2023quality, deng2021variational, meng2021magface, an_2021_pfc_iccvw}.

With the increasing size of face recognition datasets, the amount of label noise is also rising~\cite{wang2018devil}, dramatically preventing the FR from producing more effective performance. Typically, these datasets are compiled by retrieving labeled images from web search engines and cleaned using automated procedures, e.g. MS-Celeb-1M~\cite{guo2016ms}. The MS1M~\cite{guo2016ms} dataset is available in several versions: e.g. MS1Mv2, and MS1Mv3. MS1Mv2 refines the dataset by removing samples that are distant from the feature center using the model proposed by \cite{Deng_Zhou_Zafeiriou_2017}. MS1Mv3 further enhances the dataset by employing RetinaFace~\cite{deng2020retinaface} for face detection and a face recognition (FR) model to eliminate noise and duplicate samples. Another way to refrain from the impact of noise is manually labelling, however such label cleaning process requires significant human effort, which is time-consuming and resource-intensive. Therefore, it is vital to explore effective training methods for the automatic and efficient handling of label noise in large-scale datasets. 

\begin{figure}[t]
	\centering
	\includegraphics[width=0.8\columnwidth]{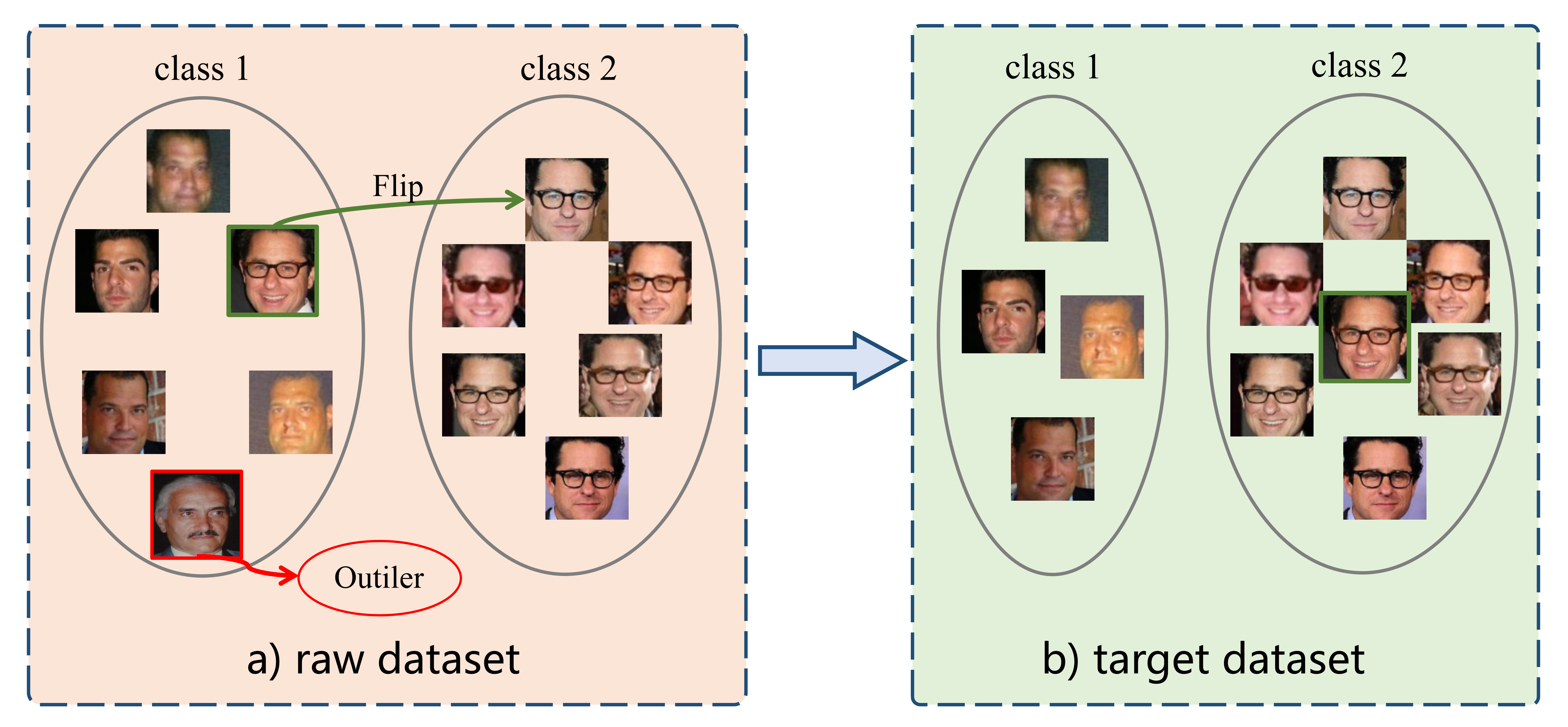}
    \caption{Noise category. a) The label noise present in the face recognition training set, which may include both closed-set(Flip) and open-set label noise(Outlier), and b) the target dataset to be obtained through noise filtering and label correction.}
	\label{fig:fig1}
\end{figure}

Label noise falls into two main categories, as detailed in Fig~\ref{fig:fig1} (a): closed-set noise and open-set noise. Closed-set noise refers to the misassignment of a person's identity to another existing identity within the dataset. In contrast, open-set noise occurs when the true/actual label of an image is absent from the dataset but is erroneously treated as an existing identity of the dataset. Studies for tackling label noise have employed two primary strategies: \textbf{noise cleaning}  and \textbf{label correction}. 
 
Recent research on \textbf{noise cleaning} predominantly adopts the seminal approaches of identifying noisy samples by adopting noise filtering techniques, subsequently these identified noisy samples are discarded or re-weighted to minimize their impact during training~\cite{hu2019noise, zhong2021dynamic, deng2020sub, wang2022rvface, wang2019co}. The above-mentioned solutions are reasonable for open-set noise scenarios, however, the majority proportion of label noise distributed in the dataset is close-set noise~\cite{wei2021open, wang2018devil}, which brings greater impact than open-set noise, thereby reducing the effectiveness of these methods when applied to real-world applications.
 
To address the problem of closed-set noise, \textbf{label correction} are adopted to assign the closed-set noise sample with their estimated label, allowing these samples to contribute to the facial recognition training without being discarded. BoundaryFace~\cite{wu2022boundaryface} pioneering focuses on detecting and correcting closed-set noise from the perspective of the decision boundary. However, this method requires FR model to obtain a certain discriminative ability in the early-stage training, but may result in inaccurate noise assessment during later-stage of training if early-stage training is not converged enough. Furthermore, this approach is dedicated to the identification and correction of clean and noisy samples, which is only effective in addressing significantly noisy samples that can be clearly identified as noise. It does not accommodate ambiguous ones that lack the requisite confidence level for definitive classification as noise. 
 
To address the abovementioned problems, we first propose a module Auxiliary Sample Cleaning (ASC) aimed at reducing the influence of noise in early-stage training. This enables the model to learn effective performance, thereby providing a discriminative ability to identify noise in the subsequent stages. Subsequently, this paper partitions the dataset into three categories: clean samples, ambiguous samples, and closed-set noise samples according to the cosine distance between the nearest negative and positive class centers. Each category is subjected to different supervision strategies. Specifically, based on the split categories, for ambiguous samples, we introduce Label Robust Fusion (LRF) method merges the prediction results of the model with the ground-truth label, aiming to maximize the use of these samples. Additionally, Smoothing Label Correction (SLC) is proposed to rectify closed-set noise samples. The main contributions of our work are as follows:

\begin{itemize}
	\item   We propose the ASC module to effectively identify and remove noisy samples during the early-stage of training, allowing the model to concentrate on effectively and efficiently learning generalized representation from clean samples. Consequently, this approach enhances model generalization and improves the model's performance in subsequent noise detection tasks.
	\item To maximize the utilization of training samples, we introduce two techniques: LRF uses a memory bank to accumulate model's prediction and fuse it with ground-truth labels, enabling more effective utilization of ambiguous samples. SLC applies an exponential smoothing strategy among the ground-truth label and corrected label to minimize the impact of incorrect corrections
	\item We conduct extensive experimentation on datasets with varying noise ratios, which achieves state-of-the-art results in well-established benchmarks, validating the effectiveness of the proposed method in handling closed-set label noise.
\end{itemize}

\section{Related Work}

\subsection{Face Recognition with Label Noise}

Researches on label noise processing methods primarily focus on cleaning label noise and correcting closed-set noisy labels. For label noise cleaning, Noise-Tolerant~\cite{hu2019noise} find that the $\theta$-distribution of training samples implicitly reflects their clean probability. Based on this finding, they propose a new paradigm for dynamically adjusting the training weights of samples. Co-Mining~\cite{wang2019co} trains two networks simultaneously, and categorizes samples into three parts: noisy, high-confidence clean, and clean samples. Based on the difference in results between the two networks, different loss functions and training strategies are used for each category of samples. SubCenter-ArcFace~\cite{deng2020sub} trains multiple sub-centers, while discarding non-dominant sub-centers and high-confidence noisy samples to enhance internal compactness and reduce the influence of noise. The noise is dynamically determined by DTDD~\cite{zhong2021dynamic} through the calculation of the cumulative average difference between the maximum cosine similarity and the label-specified cosine similarity, which is then discarded during training. RVFace~\cite{wang2022rvface} uses the OTSU algorithm to compute the noise threshold based on the $\theta$ distribution. During the training process, samples are categorized into simple, semi-hard, and ambiguous samples, with higher weights assigned to semi-hard samples.

For the correction of close-set noise labels, BoundaryFace~\cite{wu2022boundaryface} analyzes the relationship between the nearest negative and ground-truth classes from a decision boundary perspective to identify and correct closed-set noise. Part of our proposed method is further optimized based on BoundaryFace by using the smoothing technique while scaling the criteria for judging noise. The goal of our optimization is to enhance the model's capability to detect label noise and minimize the impact of misjudged noise.

\section{Methodology}

\begin{figure*}[ht]
	\centering	
        \includegraphics[width=0.8\linewidth]{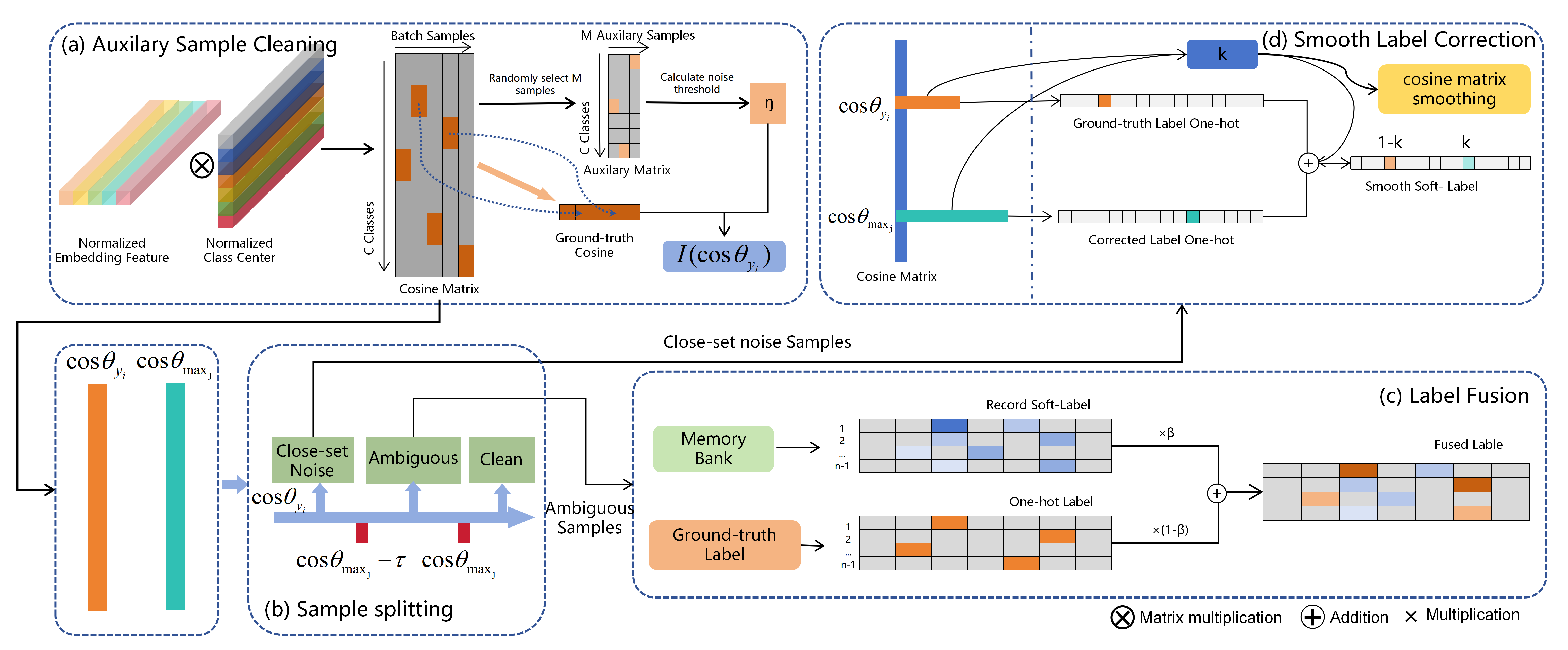}
	\caption{The overall framework. (a) Auxiliary Sample Cleaning, randomly selected samples are assigned with random label. then we compute the normalized cosine similarity between the sample and class center to be stored in the cosine matrix. It is compared by the dynamic threshold $\eta$, which is determined by the average of the cosine similarity of the Auxiliary Sample to the assigned random label. (b) Sample splitting, we dynamically split the samples into clean, ambiguous and noise samples by comparing the sample similarity to the positive class center and sample similarity to the nearest negative centre. (c) Label Fusion. We adopt a memory bank to record the accumulated model predictions as label for stabilizing the ambiguous samples training. (d) Smooth Label Correction. We opt for smoothing the label between the positive and the nearest negative class as well as the cosine similarity logits.}
	\label{fig:1102}
\end{figure*}

 Firstly, the first sub-section describes the Auxiliary Sample Cleaning (ASC) module for filtering closed-set noise during the early-stage of training. Then, we introduce how we divide samples into tree categories~(clean, ambiguous and close-set noise) in the second sub-section, and thus propose separate learning strategies for each category. Specifically, For ambiguous samples, we use a Label Fusion method namely Label Robust Fusion(LRF) to integrate stored predicted cosine logits with the labels as illustrated in the second sub-section. Finally, in the third sub-section for closed-set noise samples, we apply the smooth technique to correct the labels. The pseudocode of this paper is available at the \textit{SUPPLEMENTARY MATERIAL} Algorithm 1.

\subsection{Auxiliary Sample Cleaning}\label{sec:ASC}

Recent face recognition methods for label correction, such as BoundaryFace~\cite{wu2022boundaryface}, identify and correct label noise by analyzing the cosine similarity between sample labels and their nearest negative class, after the model has attained a certain level of recognition capability. However, during the early stages of training, when the model is still developing its learning capabilities, it remains susceptible to label noise. This early-stage noise can subsequently impair the model's ability to accurately identify label noise in the later stage of training. To address this issue, we propose an Auxiliary Sample Cleaning (ASC) strategy as illustrated in the left part of Fig~\ref{fig:1102}. This approach aims to filter out noisy samples and focus on clean samples during the early-stage training, facilitating the learning of effective facial embeddings that can be utilized for noise detection in later stage. Firstly, random $M$ samples are selected in each mini-batch and assigned with random labels $y_r$, denoted as Auxiliary Samples in this paper. Subsequently, to identify the injected Auxiliary Samples during training, we create a matrix to record the cosine similarity between the Auxiliary Samples and their randomly assigned label, which are employed to filter out the noisy samples. Specifically, the cosine similarities are calculated between each facial embedding produced by the under-training FR backbone and the weight of the linear layer representing the estimated class center. 

The average cosine similarity between samples and assigned label center in the matrix is used as a threshold, which is formulated as follows:
\begin{equation}\label{thresh}
	\eta =\frac1M\sum_{i=1}^M\cos\theta_{\gamma} + \alpha
\end{equation}where $M$ denotes the number of auxiliary samples generated in the mini-batch, $\cos\theta_{\gamma}$ denotes the cosine similarity between the samples and their assigned labels, and $\alpha$ is a parameter that stabilizes the threshold for random noise. Finally, the sample indicator can be obtained by comparing the cosine similarity between the sample and ground-truth label with the threshold calculated by the ASC:
\begin{equation}\label{Indicator}
	I(\cos\theta_{y_i}) = \begin{cases}
		0,	&	\cos\theta_{y_i}<\eta\\
		1,	&	\cos\theta_{y_i} \ge\eta\\
	\end{cases}
\end{equation}
where $\cos\theta_{y_i}$ denotes the cosine similarity between features and learned class centers corresponding to the ground-truth (actual) labels, $I(\cos\theta_{y_i})=1$ indicates the sample is clean without label perturbation, $I(\cos\theta_{y_i})=0$ suggest the sample is regarded as noisy sample. We multiply $I(\cos\theta_{y_i})=0$ by the final loss function to filter out samples identified as noisy. It is worth noting that our auxiliary sample strategy remains effective throughout the entire training process.

\subsection{Sample Splitting}
In BoundaryFace~\cite{wu2022boundaryface}, samples are only categorized as either clean or noisy. However, this method may treat noisy samples with insufficient confidence (a subset of ambiguous samples) as clean, leading to instability in the later stages of training. To address this issue, we categorize the samples into three groups: clean samples (including easy and hard sample), ambiguous samples, and closed-set noise samples, by examining the cosine similarity between the positive class and the nearest negative class, as shown in the left bottom (b) part of Fig~\ref{fig:1102}.
\begin{equation}\label{eqDis}
	d_i = \cos\theta_{max_j}-\cos\theta_{y_i}
\end{equation}where $\cos\theta_{max_j}$ denotes the cosine similarity between the sample's feature to the estimated nearest negative class center.
Specifically, samples are regarded as clean if $d_i$ is less than $0$; And samples are treated as ambiguous when $0<d_{i}<\tau$, where $\tau$ is a learned threshold for determination (discussed in Ablation Study). Finally, samples with a distance larger than $\tau$ are classified as noise samples.
 
\subsection{Ambiguous Sample Label Fusion}\label{RLF}
The ambiguous samples are distributed close to the decision boundary, thus the actual label of these samples cannot be explicitly applied for training. To appropriately tackle ambiguous samples, inspired by SELC~\cite{lu2022selc}, which update the original noisy labels through ensemble predictions, we propose the Label Robust Fusion (LRF) module, illustrated in the right bottom(c) part of Fig~\ref{fig:1102}. It effectively fuses the model's cosine prediction logits and labels of ambiguous samples, aiming at adopting the accumulated prediction at training for stabilizing the learning of ambiguous samples. 
 
Specifically, we use a memory bank to store the model's prediction results during training, by recording the maximum cosine logits from the model’s prediction results, representing the distance between the sample features and the learned class center. If a cosine logits for a class has already been recorded, it is updated according to the formula~\ref{lfcos}. Otherwise, the new logits $\cos(\theta_{max_j})$ is stored directly.
\begin{equation}\label{lfcos}
	\cos\theta_{p} = (1-\beta)*\cos(\theta_{p}) + \beta*\cos(\theta_{max_j})
\end{equation}
where $\cos\theta_{p}$ represents the value of the recorded cosine logits, $\cos\theta_{max_j}$ represents the value of the maximum cosine similarity between the sample features and the learned weights at the current prediction, and $\beta$ is a balancing parameter.

During the training process, we continuously update the model's max cosine prediction logits in the memory bank. It is worth noting that a sample may have multiple different logits, we normalize these logits in order to generate soft labels. Subsequently, ground-truth one-hot encoding is adopted for fusion with the generated soft labels, as depicted in Fig \ref{fig:1102} (Label Fusion(c)) and the fusion formula is as follows. 

\begin{equation}\label{lflabel}
	\boldsymbol{q} _{r}= \beta\boldsymbol{p} + (1-\beta) \boldsymbol{q}_{y_i}
\end{equation}where $\boldsymbol{p}$ and $\boldsymbol{q}_{y_i}$ represent soft labels generated by the stored cosine logits and the one-hot ground-truth label, respectively, and $\beta$ is the same as Eq. \ref{lfcos}.

Ambiguous samples could be either clean or noisy. Therefore, if the samples are actually clean, the preserved cosine logits’ related category often aligns with the ground-truth label. In such cases, the fused label aligns with the class existing in the dataset. Conversely, if they are closed-set noise samples, the stored cosine logits' related category often represents the real label. By fusing labels together, we enhance the robustness of our model.

\subsection{Smoothing Label Correction}\label{SLC}

In our method, we utilize the disparity between the cosine similarity of the nearest negative class center and positive class center for each sample (using Eq. \ref{eqDis}, denote as $d_i$) as a metric for assessing noise. 

Although BoundaryFace~\cite{wu2022boundaryface} has demonstrated a commendable ability to detect and correct label noise, an analysis of its label detection formula reveals certain limitations: $if \max\left\{ \cos(\theta_k+m)~for~all~k\ne y_i\right\}-\cos{\theta_{y_i}} > 0: y_i=k$, where the $m$ is the margin in the loss function. According to the formula, the smaller $\theta_{y_i}$ is, the more stringent the criteria for determining noise will be. This indicates that the method imposes stricter criteria on samples that cannot be accurately predicted, resulting in label correction only for samples with high confidence. Consequently, this can leave some label noise uncorrected during training. To address this issue and maximize label noise correction while maintaining model stability, we propose the Smoothing Label Correction (SLC) module, shown in the top right part(d) of Fig~\ref{fig:1102}. A fixed, small threshold $\tau$ is used as the criterion for identifying closed-set noise. While lowering this threshold can relax the noise detection criteria, it also increases the risk of misjudgment. To address this, we propose a smoothing method that relaxes the criteria while reducing the impact of misjudgment errors. 
For hard sample mininig, MV-Softmax~\cite{wang2020mis} is used as our baseline, which is formulated as follows.
\begin{equation}\label{hardraw}
	N_1(t,\cos\theta_j)=\begin{cases}
		\cos\theta_j  &   T(\cos\theta_{y_i})\ge \cos\theta_j \\
		\cos\theta_j+t & T(\cos\theta_{y_i}) < \cos\theta_j
	\end{cases}
\end{equation}where $T(\cos\theta_{y_i})=\cos(\theta_{y_i}+m)$ follow MV-softmax to use the $m$ in ArcFace~\cite{deng2019arcface}.

To balance the ratio between the two parts of the label and the nearest negative class, we compute a dynamic balancing parameter $k = \text{Sigmoid}(10 \cdot d_i)$, where the $\text{Sigmoid}$ function is used to keep the ratio between 0 and 1, and multiplying the distance by 10 is used to expand the influence factor of the cosine-valued distance. Due to the label correction, we need to perform a smoothing weight adjustment for noising sample mining. The cosine matrix smoothing process is:
\begin{equation}\label{hardcorrect}
	N_2(t,\cos\theta_j)=(1-k) N_1(t, \cos\theta_{1}) 
	+k N_1(t,\cos\theta_{2})
\end{equation}where $N_1(t, \cos\theta_{1})$ indicates the hard samples that are computed based on the given labels, and $N_1(t,\cos\theta_{2})$ indicates the hard samples computed based on the corrected label which is refined by Eq.~\ref{hardcorrectlb}. After applying a smoothing operation to the noise sample mining, and equally importantly, we employ a similar approach to smooth the labels by transforming hard labels into soft labels. The formula for label correction by using the smoothing technique is as follows:
\begin{equation}\label{hardcorrectlb}
	\boldsymbol{q}_{s}=(1-k)\boldsymbol{q}_{y_i} + k \boldsymbol{q}_{y_j}
\end{equation}where $\boldsymbol{q}_{y_i}$ and $\boldsymbol{q}_{y_j}$ represent the one-hot encoding of the true label and the nearest negative label for closed-set noise, respectively.

By dividing the samples into clean, closed-set noise and ambiguous samples, for which we use different strategies to enhance the resilience of the network model against closed-set noise, the overall framework is referred to as RepFace. which can be summarised as follows:
\begin{equation}\label{prog}
	\mathcal{L}_1 = 	 \begin{cases}
		-I(\cos\theta_{y_i})\cdot \boldsymbol{q}_{s} \cdot logP_{y_j} & d_i>\tau\\
		-I(\cos\theta_{y_i})\cdot \boldsymbol{q}_{r}\cdot logP_{y_i} & \tau \ge d_i\ge 0\\
		-I(\cos\theta_{y_i})\cdot \boldsymbol{q}_{y_i}\cdot logP_{y_i} & others\\
	\end{cases} 
\end{equation}
\begin{displaymath}
	P_{y_j}= \frac{e^{s\cdot T(\cos\theta_{y_i})}}{e^{s\cdot T(\cos\theta_{y_i})}+\sum\limits_{j=1,j\ne i}^Ce^{s\cdot N_2(t,\cos\theta_j)}}
\end{displaymath}
\begin{displaymath}
	P_{y_i} =  \frac{e^{s\cdot T(\cos\theta_{y_i})}}{e^{s\cdot T(\cos\theta_{y_i})}+\sum\limits_{j=1,j\ne i}^{C}e^{s\cdot N_1(t,\cos\theta_j)}}
\end{displaymath}

\section{Experiments}

\subsection{Experiment Settings}

\textbf{Datasets.} Our training set is the widely-used MS1MV2~\cite{guo2016ms,deng2019arcface} and CASIA-WebFace~\cite{yi2014learning}, which contains approximately 5.8M images of 85K individuals and approximately 0.5M images of 10K individuals, respectively. To simulate label noise within the dataset, we manually synthesize closed-set noise for constructing a noisy dataset. Specifically, we randomly select a certain percentage of samples from each person in the dataset and assign them with random labels. According to~\cite{wang2018devil}, CASIA-WebFace contains 9.3\%-13\% noise, therefore, we artificially synthesize noisy datasets with 10\% and 20\% proportions to reproduce similar noise volume to CASIA-WebFace. Finally, we perform experiments using synthetic datasets and clean datasets. Our test datasets are several popular benchmarks in face recognition including \textit{LFW}~\cite{huang2008labeled}, \textit{SSLFW}~\cite{deng2017fine}, \textit{CALFW}~\cite{zheng2017cross}, CPLFW\cite{zheng2018cross}, \textit{AgeDB}~\cite{moschoglou2017agedb}, \textit{CFP-FP}~\cite{sengupta2016frontal}, and \textit{RFW}~\cite{wang2018racial}, \textit{IJB-B}~\cite{whitelam2017iarpa}, and \textit{IJB-C}~\cite{maze2018iarpa}. The metrics for evaluating each dataset are, TAR@FAR=1e-5 and 1e-4 for evaluation on \textit{IJB-B}, and \textit{IJB-C}; 1:1 verification accuracy are adopted for LFW, SSLFW, CALFW, AgeDB, CFP-FP, CPLFW and RFW.

\noindent\textbf{Training details.} We follow the ArcFace~\cite{deng2019arcface} to crop and resize faces in a size of  112$\times$112. IR-ResNet50~\cite{duta2020improved} is the backbone with a feature output dimension of 512. The models are optimized by the typical stochastic gradient descent(SGD) algorithm with a momentum of 0.9 and weight decay of 5e-4. The learning rate starts at 0.1 and is divided by 10 at the 9th, 18th, and 26-th epochs, with a total of 30 training epochs on CIAIA-WebFace, while the learning rate is divided by 10 at the 10th, 18th, and 22-nd epochs, with totally 24 training epochs on MS1MV2~\cite{deng2019arcface}. We set the scale hyperparameters to $s=64$, margin is $m=0.5$.

\subsection{Ablation Studies}
A number of studies are carried out to demonstrate the effectiveness of each part of our proposed RepFace. The noise ratio employed in this study is 20\% unless specified. The default backbone is IR-50~\cite{duta2020improved}. Due to page limitation, we put the discussion of $M$, the effectiveness of our method on another advanced loss and the impact of open-set noise in the \textit{SUPPLEMENTARY MATERIAL} .

\subsubsection{Hyperparameter Analysis}\label{Sec:hyperparameter}
\begin{figure}[b]
	\centering
	\includegraphics[width=0.8\columnwidth]{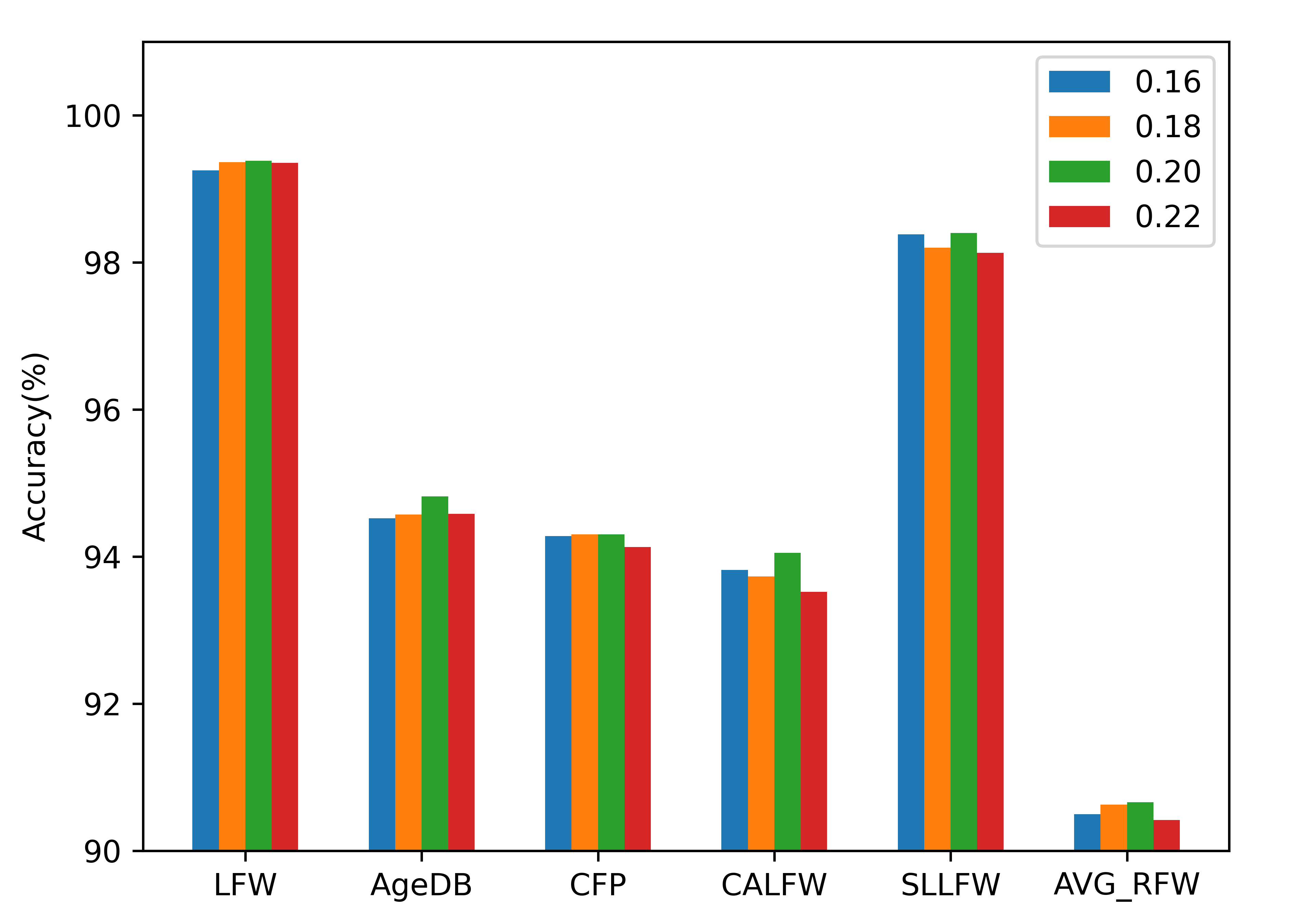}
	\caption{\small Test results on\textit{ LFW, AgeDB, CFP-FP, CALFW, and SLLFW} with different hyperparametric $\tau$.}
	\label{fig:t}
\end{figure}

\paragraph{Impact of $\tau$}
In our comprehensive study, we initially analyze the critical threshold value $\tau$, which is used to differentiate and assess closed-set noise by thoroughly comparing the threshold utilized by the Boundaryface algorithm. Notably, the noise judgment in BoundaryFace imposes a larger restriction in determining noise when the angle is significantly small. In our innovative approach to enhance closed-set noise detection, we meticulously refine and narrow judgment thresholds. Intuitively, we set $\tau$ to four distinct values: 0.16, 0.18, 0.20, and 0.22 as given in Fig~\ref{fig:t}. Experiments on datasets with 10\% closed-set noise enabled rigorous evaluation. Among the four $\tau$ values tested, we can find that, if $\tau$ set too large, it will reduce the number of detected label noise instances, while a smaller value increases the risk of incorrect detections. Based on these findings, we decide to set $\tau=0.2$ for all subsequent experiments. 

\paragraph{Impact of $\alpha$}
In the first sub-section of Methodology, we initially screen the samples by generating auxiliary noise samples, and use Equation~\ref{thresh} to calculate the threshold, where $\alpha$ to stabilize the threshold calculation. We use 0.03, 0.05 and 0.07 as alternatives to conduct experiments. From the experimental results Table~\ref{tab:freq}, we can find that in each verification dataset, when $\alpha$ is 0.05, our method achieves the best effect. In addition, the performance of our method will degrade when $\alpha$ is greater than or less than 0.05, so we adopt the value of $\alpha$ to be 0.05 in subsequent experiments.

\begin{table}[h]
    \centering
        \small
		\begin{tabular}{cccccc}
			\hline
			$\alpha$ &\textit{LFW} & \textit{AgeDB} &\textit{CFP}&\textit{CPLFW}&\textit{RFW(Avg.)} \\
			\hline
			0.03 & 99.23 &	94.23 &	93.31 &88.20 & 89.59\\
			0.05 & \textbf{99.40} &	\textbf{94.43} &	93.44&	\textbf{88.57}  &	\textbf{90.10}\\
			0.07 & 99.22 &	94.32 &	\textbf{93.89}  &	88.25&	89.72\\
			\hline
		\end{tabular}
        \caption{Verification performance(\%) with different $\alpha$.}\label{tab:freq}

\end{table}

\paragraph{Impact of $\beta$}
Based on our analysis above, in the critical matter of selecting the balance parameter $\beta$, we have chosen to use values of 0.85, 0.9, and 0.95, for our experimentation. A meticulous examination of the results, as detailed in Table \ref{bb}, reveals a clear and definitive trend: our RepFace demonstrates its superiority in performance when the value of $\beta$ is set to 0.9. Consequently, for the remainder of this study, the value of $\beta$ has been firmly opted for at 0.9, providing a solid foundation for further exploration and analysis.
\begin{table}[h]
	\centering
 	\small
	\begin{tabular}{cccc}
		\hline
		Methods& \textit{LFW} & \textit{AgeDB} & \textit{CFP} \\ 
		\hline
		$\beta=0.95$ & 99.33 & 94.32 & 93.30\\ 
		$\beta=0.9$ & \textbf{99.40} & \textbf{94.43} & \textbf{93.44}\\ 
		$\beta=0.85$ & 99.33 & 94.37 & 93.40\\
		\hline
	\end{tabular}
  \caption{Verification performance(\%) of different $\beta$.}\label{bb}
 
\end{table}

\subsubsection{Impact of each module}
\paragraph{Effect of each module} To rigorously ascertain the efficacy of each individually proposed module, a set of experiments are executed. This involves incorporating each module sequentially into the baseline MV-Softmax. Initially, MV-Softmax is trained solely on the noisy dataset. Subsequently, we augment this baseline with the ASC module, followed by the SLC and LRF modules, each added sequentially. Table \ref{ablation} serves to elucidate the distinct contributions made by each module.
\begin{table}[h]
	\begin{center}
		\small
		\begin{tabular}{cccccc}
			\hline
			Methods& \textit{LFW} & \textit{AgeDB} & \textit{CALFW}& \textit{RFW(Avg.)} \\ 
			\hline
			MV-Softmax & 99.12 & 92.58 &92.45&88.00\\
			ASC &99.13 &92.87 & 92.93&88.45\\ 
			ASC+SLC & 99.10 & 93.83& 93.12&88.78\\
            ASC+SLC+LRF & \textbf{99.40} &\textbf{94.43} & \textbf{93.50} &\textbf{90.10}\\
            \hline
		\end{tabular}
 \caption{Validation performance(\%) of progressively adding ASC,SLC and LRF to baseline MV-softmax.}\label{ablation}
	\end{center}
\end{table}

It's worth noting that the performance of our models does not show a substantial improvement with just the ASC and SLC modules included. However, there is a marked and significant enhancement in the performance of our trained models once all three modules are assimilated into the proposed framework, the RepFace. 

\paragraph{Effect of sample and label smoothing in SLC}

Herein, we ablate the choice of smoothing option described from the fourth sub-section in Methodology. Specifically, for identified samples are provided three options:(a) directly assign the nearest negative class's label to the sample; (b) correct the label with smoothing technology; (c) incorporate the cosine matrix smoothing approach while assigning the label as (a); And finally, combine the (b) and (c) as the complete SLC module provided by Section~\ref{SLC}. The results are available at Tabel~\ref{ablation_slc}, revealing that (b) and (c) attribute the training performance under noisy data, and (d) indeed validate the effectiveness of the SLC. \frenchspacing

\begin{table}[h]
\centering
\small
\setlength{\tabcolsep}{0.5mm}{
\begin{tabular}{ccccc}
\hline
Methods & \textit{LFW} & \textit{AgeDB} & \textit{CALFW} & \textit{RFW(Avg.)} \\ \hline
(a) directly assign    & 99.32 & 93.00 & 93.03 & 88.30\\
(b) label smoothing    & 99.25 & 93.60 & 93.05 & 92.89\\
(c) cosine matrix smoothing     & 99.33 & 93.90  & 93.05 & 93.11 \\
(d) SLC    & \textbf{99.40} & \textbf{94.43}  & \textbf{93.50}  & \textbf{93.50}  \\ \hline
\end{tabular}
\caption{Validation performance(\%) of effective of smoothing technology in SLC.}\label{ablation_slc}
}
\end{table}

\begin{table*}[ht]
        \centering
	\small
		\begin{tabular}{c|c|cccccc|cccc}
			\hline
			\multirow{2}{*}{Ratio} & \multirow{2}{*}{Method} & \multirow{2}{*}{\textit{LFW}} & \multirow{2}{*}{\textit{AgeDB}} & \multirow{2}{*}{\textit{CFP}} & \multirow{2}{*}{\textit{CALFW}} & \multirow{2}{*}{\textit{SLLFW}} & \multirow{2}{*}{\textit{Avg.}} & \multicolumn{4}{c}{\textit{RFW}} \\ \cline{9-12} 
			&&&&&&&&Asian & Caucasian & Indian & Afican \\
			\hline
			\multirow{8}{*}{0\%}	
			&ArcFace& 99.25 & 94.20 & 94.70 & 93.37  & 98.00&95.90  & 86.83 & 93.22 & 90.10 & 87.68\\
			&MV-Softmax& 99.37 & 93.80 & 94.76 & 93.07  & 98.17&95.83  & 87.23 & 93.67 & 89.95 & 87.88\\
			&Curricular& 99.28 & 94.48 & 94.37 & 93.25& 98.22 & 95.92 & 87.03 & 93.57 & 90.67 & 88.53\\
			&RVFace&99.42 &94.40&95.01&93.40&98.25&96.10 &87.75&93.75&91.03&88.27\\
			&BoundaryFace & 99.33 & 94.05 &\textbf{ 95.26 }& 93.43 & 97.98&96.01  & 87.50 & 94.00 & 90.55 & 87.72\\
			&AdaFace & \textbf{99.50} &	94.63 	& 94.91 & 	93.58 &	98.32&96.19  &	87.20& 	94.45 &	90.52 &	87.83 \\
			&RepFace(ours) & 99.47 & \textbf{94.77 }& 94.71 & \textbf{93.82} & \textbf{98.53}&\textbf{96.26}  & \textbf{88.42 }& \textbf{94.87} &\textbf{ 91.72} &\textbf{ 89.58}\\
			\hline
			\multirow{8}*{10\%}	
			&ArcFace & 99.22 & 93.40 & 93.29 & 92.90 & 97.45 & 95.25  & 85.87 & 92.42 & 89.82 & 86.23\\
			&MV-Softmax & 99.32 & 93.97 & 93.81 & 93.25 & 98.00 & 95.67  & 86.72 & 93.22 & 90.13 & 87.38\\
			&Curricular& 99.30 & 93.40 & 93.70 & 93.03 & 97.85 & 95.46 & 85.93 & 92.78 & 89.93 & 87.05\\
			&RVFace &99.27 &94.17&94.47&93.43&97.73& 95.81   &  87.30&93.42&90.60&87.63\\
			&BoundaryFace  & 99.25 & 94.33 & \textbf{94.73} & 93.50 & 97.98 &95.96 & 87.47 & 93.18 & 90.15 & 87.53\\
			&AdaFace &99.25 &	93.82 &	94.07 &	93.02 &	97.85 &95.60&	86.22 &	93.27 &	90.20 & 87.32 \\
			&RepFace(ours) & \textbf{99.33} & \textbf{94.78} & 94.30 & \textbf{94.05}  & \textbf{98.30}& \textbf{96.15} &\textbf{87.58} & \textbf{94.50} & \textbf{91.48} & \textbf{88.95 }\\
			\hline
			\multirow{8}*{20\%}
			&ArcFace & 98.73 & 92.30 & 91.79 & 92.48  & 97.10 & 94.48 &85.18 & 91.53 & 89.20 & 84.45 \\
			&MV-Softmax & 99.12 & 92.58 & 92.34 & 92.45 & 97.15 &94.73 & 85.68 & 91.83 & 89.50 & 84.97 \\
			&Curricular& 99.02 & 91.35 & 91.56 & 92.25 & 96.52 & 94.14 & 84.48 & 90.82 & 88.52 & 84.38\\
			& RVFace &99.10&93.60&93.86&93.10 & 97.62 & 95.45 & 86.15&92.80&89.92&87.08\\
			&BoundaryFace  & 99.38 & 94.22 & \textbf{93.89} & 93.40 & 97.90 &\textbf{95.76} & 86.23 & 93.22 & 90.00 & 87.27\\
			&AdaFace & 98.97 &	92.10 &	92.26 &	92.57&	97.15 & 94.61 &	84.98 &	91.17 &	88.73 &	84.57 \\
			&RepFace(ours) & \textbf{99.40} &\textbf{ 94.43} & 93.44 & \textbf{93.50 }& \textbf{98.03} & \textbf{95.76} &\textbf{87.25} & \textbf{93.97 }& \textbf{90.83} & \textbf{88.35 }\\

			\hline
		\end{tabular}

  \caption{Verification performance (\%) of different methods trained on CASIA-WebFace.} \label{tab:cap}
\end{table*}

\begin{table*}[h]
	\centering
	\small
	\begin{tabular}{c|c|cccccc|cc|cc}
		\hline
		\multirow{2}{*}{Ratio}&\multirow{2}{*}{Method} & \multirow{2}{*}{LFW}& \multirow{2}{*}{AgeDB-30}&\multirow{2}{*}{CFP-FP}&\multirow{2}{*}{CALFW}&\multirow{2}{*}{SLLFW}&\multirow{2}{*}{Avg.} & \multicolumn{2}{c}{IJB-B} & \multicolumn{2}{c}{IJB-C}\\\cline{9-12} 
		&&&&&&&&1.E-05 &0.0001 & 1.E-05 & 0.0001 \\ \cline{1-12} 
		\multirow{7}{*}{0\%}   & ArcFace &99.75&97.80 &	95.21 &95.90 &	99.53 &	97.64  &	88.94 &	94.31  &	93.58 &	95.80  \\
		& MV-Softmax & 99.76 &	\textbf{97.82} &	95.34 & \textbf{95.92} &	99.37 &	97.64   &	87.60 &	94.11  &	93.28 &	95.68  \\
		& Curricular & 99.71 &	97.80&	95.37&	95.78&	99.45&	97.60&	87.93&	94.25&	93.21&	95.75\\
        & BoundaryFace & 99.63 &	\textbf{97.82} &	95.46 & 95.75 &	99.32 &	97.60   &88.69 &	94.20  &	93.36 &	95.73 \\
		& RVFace & 99.78 &	97.77 &	95.74 & 95.88 	& 99.53 &	97.74   &	86.21 	& 93.62  &	91.68 &	94.99 \\
  		& AdaFace &99.75 &	97.81 &	95.84 &	95.82 &	\textbf{99.60} &	97.76 &	87.22 &	94.20  &	92.82 &	95.66  \\
		& RepFace(ours) & \textbf{99.79} &	\textbf{97.82} &	\textbf{95.86} &  \textbf{95.92} &	99.48 &	\textbf{97.77}  &	\textbf{88.98} &	\textbf{94.67} &	\textbf{94.01} &	\textbf{96.14}  \\ \hline
		\multirow{7}{*}{20\%}  
        & ArcFace & 99.65 &	97.65 &	95.05 & 95.97 &	99.48 &	97.56  &	80.13 	& 92.71  &	88.41 &	94.39  \\
		& MV-Softmax &99.64&	97.37&	95.20&	95.70&	99.34&	97.45&	80.30&	92.79	&88.67	&94.65\\
		& Curricular & 99.65&	97.55&	95.22&	95.68&	99.30&	97.48&	81.40&	92.92&	89.00&	94.89\\
		& BoundaryFace & 99.68 &	\textbf{97.69} &	\textbf{95.35} &	95.97 &	99.45 &	97.63 &	84.49 &	93.11  &	90.79 &	94.70 \\
		& RVFace & 99.68 &	97.66 &	95.26& 95.75 &	\textbf{99.50} &	97.57 &	84.09 &	92.93  &	90.60 &	94.43 \\
         & AdaFace& 99.65 &	97.27 &	93.91 & 95.73 &	99.37 &	97.19  &	75.40 &	90.31  &	83.19 &	92.27 \\
		& RepFace(ours) & \textbf{99.70} &	97.68 &\textbf{95.35} &\textbf{96.10} &\textbf{	99.50} &	\textbf{97.67} &	\textbf{84.49} &\textbf{93.12}  &\textbf{90.91} &\textbf{94.77} \\ \hline
	\end{tabular}
    \caption{Evaluation outcomes for MS1Mv2 Raw Dataset and 20\% Closed-Set Noise-Infused Synthetic Dataset.}\label{ms1mv2}
\end{table*}
\subsection{Comparisons with SOTA Methods}
In this section, we evaluate our innovatively designed RepFace framework on two synthetic, closed-set noise datasets and the raw CASIA-WebFace dataset. These datasets are capable of modelling the label noise present in realistic applications where the noise ratio varies according to different collection environments. Open-set noise is not considered as the prominent disturbance for effective FR training is closed-set noise. Table~\ref{tab:cap} illustrates the comparison between our methodology with several state-of-the-art (SOTA) alternatives, including ArcFace, MV-Softmax, CurricularFace, RVFace, AdaFace, and BoundaryFace. Our proposed RepFace achieves 96.26 on \textit{avg}, 88.42 on Asian, 94.87 on  Caucasian, 91.72 on Indian and 89.58 on Afican. In terms of injecting noise, experiment, our method improves the previous BoundaryFace by 0.19 on the \textit{avg}, and the gap is also large when using the heavy noisy ratio 20\%.  Another interesting conclusion can be drawn by comparing the gap between the real and 20\% noisy dataset, for instance, AdaFace decreases by 1.58, however RepFace drops by 0.5, demonstrating the robustness to noise of our method. 

Observing suggests that our methodology demonstrates superior performance compared to the state-of-the-art (SOTA) method across both synthetic closed-set noise sets and the primitive CASIA-WebFace dataset, validating feasibility and effectiveness on synthetic noisy datasets.

\subsection{Generalizability Evaluation}

To validate the generalizability of our method, we conduct further experiments of training on the MS1Mv2 dataset. Specifically, our experiments are divided into two parts: training on the original MS1Mv2 dataset and training on a synthetic dataset with 20\% closed-set noise. We make evaluation experiments on the mainstream verification datasets, IJB-B, and IJB-C datasets.

For the original dataset, we compare our method with mainstream techniques for handling hard samples and label noise.
For the noisy dataset, we compare our method with leading label noise processing approaches. The results of model trained on the original dataset are illustrated in Table~\ref{ms1mv2}, we achieved results comparable to state-of-the-art (SOTA) methods on LFW, AgeDB-30, CFP-FP and CALFW, with our average performance being the best, surpassing the baseline MV-Softmax by 0.13. Similarly, our method demonstrates the highest performance across all metrics in the IJB-B and IJB-C datasets. These findings indicate that our method consistently outperforms other approaches when trained on the original dataset.

Although our results are comparable to state-of-the-art (SOTA) methods on the original dataset, the advantages of our method become more pronounced when the noise level is increased to 20\%. In the experiments with the synthetic 20\% closed-set noise dataset, our method exhibits better performance compared to previous label noise processing methods. Our method achieves the best results in both average validation sets and IJB-B/IJB-C test sets, especially in the IJB-C dataset where TAR@FAR=1e-4, surpassing BoundaryFace and RVFace by 0.07\% and 0.34\%, respectively. These results confirm that our method remains effective in large-scale datasets, thereby validating its robustness and generalizability.

\section{Conclusion}
This paper proposes a novel closed-set noise training framework, namely RepFace. Firstly, we use generated auxiliary closed-set noise samples to simulate closed-set noise sample, and propose an ASC model to identify and discard noisy samples, especially during the early-stage training of the model. Subsequently, we partition the datasets into three parts (clean, ambiguous, and noise) based on the cosine similarity difference between the nearest negative and positive classes. We introduce a label robust fusion module by injecting a memory bank solution to accumulate model prediction for effectively handling these ambiguous samples. Moreover, we apply smooth label correction using the smooth parameter $k$ to correct closed-set noise samples. This framework minimizes the impact of label noise on the model during early training and maximizes utilization of ambiguous and noise samples during post-training. Extensive experiments validate our method's efficacy and achieve state-of-the-art results.

\section{Acknowledgments}
This work is partially supported by National Natural Science Foundation of China (62376231), Sichuan Science and Technology Program (24NSFSC1070), Tangshan Basic Research Science and Technology Program (23130230E), Tangshan Lunan District Science and Technology Program. Fundamental Research Funds for the Central Universities (2682023ZDPY001).

\bibliography{reference.bib}

\clearpage

\twocolumn[
\begin{@twocolumnfalse}
	\begin{center}
		{\LARGE \textbf{Supplementary Material}\\[1em]} 
	\end{center}
\end{@twocolumnfalse}
]
This is the supplementary material for the paper \textbf{RepFace: Refining Closed-Set Noise with Progressive Label Correction for Face Recognition}. The code will be made publicly available in the near future.

\section{Addition to Experiments}
\subsection{Impact of $M$}
In the first sub-section of our Methodology, we introduce the parameter $M$. For our experiments, with a batch size of 256, we tested values of 24, 32, 40, and 48 on CASIA-WebFace dataset with 20\% synthesized noise. The results, presented in Table~\ref{ablation_m}, show that the impact of $M$ on our proposed method is negligible. We set $M$ to 32 in our final implementation.
\begin{table}[h]
	\centering
	\small
	\begin{tabular}{ccccc}
		\hline
		M  & LFW            & AgeDB-30       & CALFW          & RFW(Avg.)      \\ \hline
		24 & 99.39          & \textbf{94.43} & \textbf{93.50} & 90.10          \\
		32 & \textbf{99.40} & \textbf{94.43} & \textbf{93.50} & \textbf{90.11} \\
		40 & 99.38          & 94.42          & \textbf{93.50} & 90.10          \\
		48 & 99.39          & \textbf{94.43} & 93.49          & 90.07          \\ \hline
	\end{tabular}
	\caption{Validation performance(\%) of different $M$.}\label{ablation_m}
\end{table}

\subsection{Combine with advanced hard-sample mining loss}
Our method also achieves strong performance across other baselines. Herein we examine if an advanced loss is of improvement for our proposed method. By incorporating RepFace into AdaFace and testing on a dataset with 20\% closed-set noise given in Tabel~\ref{ablation_ada}, we found that our approach remains effective within AdaFace. These results further validate the effectiveness and generalizability of our method.
\begin{table}[h]
	\centering
	\small
	\setlength{\tabcolsep}{1mm}{
		\begin{tabular}{cccccc}
			\hline
			Methods& \textit{LFW} & \textit{AgeDB} & \textit{CALFW}& \textit{RFW(Avg.)} \\ 
			\hline
			MV-Softmax & 99.12 & 92.58 &92.45&88.00\\
			MV-Softmax + RepFace&\textbf{99.40}&\textbf{94.43}&\textbf{93.50}&\textbf{90.10}\\
			\hline
			AdaFace& 98.97 &92.10 &92.57 & 87.36\\
			AdaFace + RepFace&\textbf{99.33} &	\textbf{94.48} & \textbf{93.17} &\textbf{89.62} \\
			\hline
		\end{tabular}
	}
	\caption{Verification performance(\%) of our method on different hard sample mining loss.}\label{ablation_ada}
	
\end{table}

\subsection{Evaluation on RFW With MS1M Dataset}

To further improve the effectiveness of our method, we conducted additional tests on the model trained with MS1Mv2 (including raw datasets and synthetic 20\% close-set noise dataset). The test results on RFW are shown in Table~\ref{ms1mv2rfw}. From these results, we can see that our method achieves the best performance across multiple ethnic groups, with significant improvements for groups with fewer samples in the training dataset, such as Asians and Africans. These experimental results demonstrate that our method has strong generalization capabilities, enabling more accurate feature extraction for underrepresented groups, thereby validating the effectiveness and robustness of our approach.
\begin{table}[h]
	\centering
	\small
	\begin{tabular}{c|c|cccc}
		\hline
		Ratio & Method & Asian & Cauc& Indi & Afri\\\cline{1-6}
		\multirow{5}{*}{0\%}   
		& ArcFace &	97.00&	\textbf{98.85}	&97.60&	96.98 \\
		& MV-Softmax & 96.30&	98.70&	97.28&	96.73  \\
		& Curricular & 96.30&	98.73&	97.43&	96.88\\
		& BoundaryFace & 96.52&	98.57&	97.25&	96.68 \\
		& RepFace(ours) &\textbf{97.28}&	\textbf{98.85}&\textbf{	97.87}&	\textbf{97.53} \\ \hline
		\multirow{5}{*}{20\%} 
		& ArcFace & 96.57&	98.77&	97.1&	97.50  \\
		& MV-Softmax &95.99&	98.20&	96.98&	96.42\\
		& Curricular &96.03	&98.33&	97.12&	96.50\\
		& BoundaryFace & 96.63&	\textbf{98.83}&	97.3&	97.47 \\
		& RepFace(ours) & \textbf{97.12}&	\textbf{98.83}&\textbf{	97.82}&	\textbf{97.67 }\\ \hline
	\end{tabular}
\caption{Evaluation outcomes for MS1Mv2 Raw Dataset and 20\% Closed-Set Noise-Infused Synthetic Dataset.}\label{ms1mv2rfw}
\end{table}

\subsection{Evaluation in datasets containing both open and closed set noise}
To further validate the effectiveness of our proposed RepFace framework in open-set noise scenarios, we utilize CASIA-WebFace as the original dataset and introduce both closed-set and open-set noise, similar to our previous approach. We synthesize noisy datasets with 10\% open-set and 10\% closed-set noise. We validate the effectiveness of our method on these synthesized datasets. Table~\ref{openclose} demonstrates that while the performance of each method declines when dealing with datasets containing open-set noise, our proposed RepFace framework continues to achieve high performance, which verifies that the RepFace framework is also robust to open-set noise.
\begin{table}[h]
	\small
	\centering
	\begin{tabular}{cccccc}
		\hline
		Methods& \textit{LFW} & \textit{CFP} & \textit{CALFW}& \textit{SLLFW} \\ 
		\hline
		ArcFace & 95.58 & 90.66 & 87.33 & 91.78\\
		CurricularFace & 97.78 &91.27 & 88.92 & 94.87\\
		BoundaryFace & 98.55 & 91.60& 90.83 & 96.20\\
		RepFace(ours) & \textbf{98.65} & \textbf{91.82} & \textbf{91.92}  & \textbf{96.67}\\
		\hline
	\end{tabular}
	\caption{Validation performance(\%) on 10\% Open-set and 10\% Closed-set Dataset.}\label{openclose}
\end{table}

\subsection{Convergence}

\begin{figure}[h]
	\centering
	\includegraphics[width=1\linewidth]{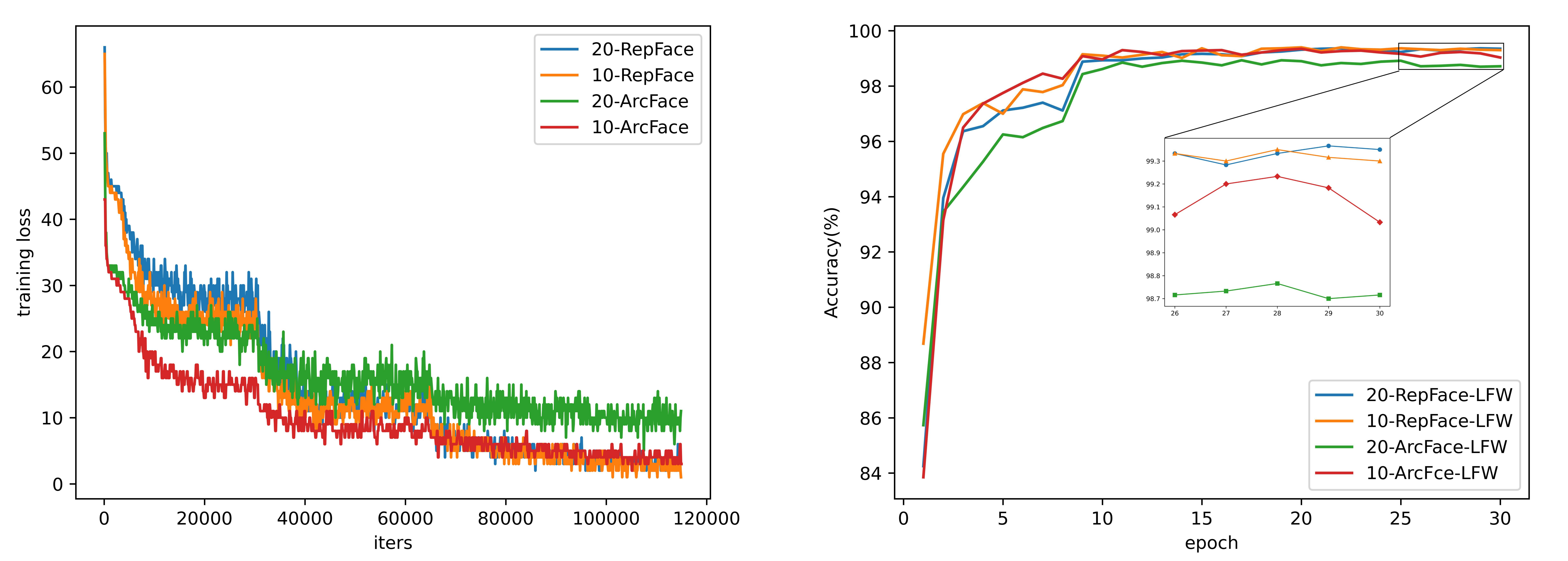}
	\caption{Loss and \textit{LFW}~\cite{huang2008labeled} accuracy of ArcFace~\cite{deng2019arcface} and the proposed method in this paper trained on 10\% and 20\% closed-set noise CASIA-WebFace~\cite{yi2014learning} dataset.}
	\label{loss_acc}
\end{figure}

It can be seen from Figure \ref{loss_acc} that our proposed method can converge more stably in datasets with different noise ratios, and at the same time demonstrate competitive results in LFW datasets.

\section{Discussion of Label Correction}

\begin{figure}[h]
	\centering	\includegraphics[width=0.8\linewidth]{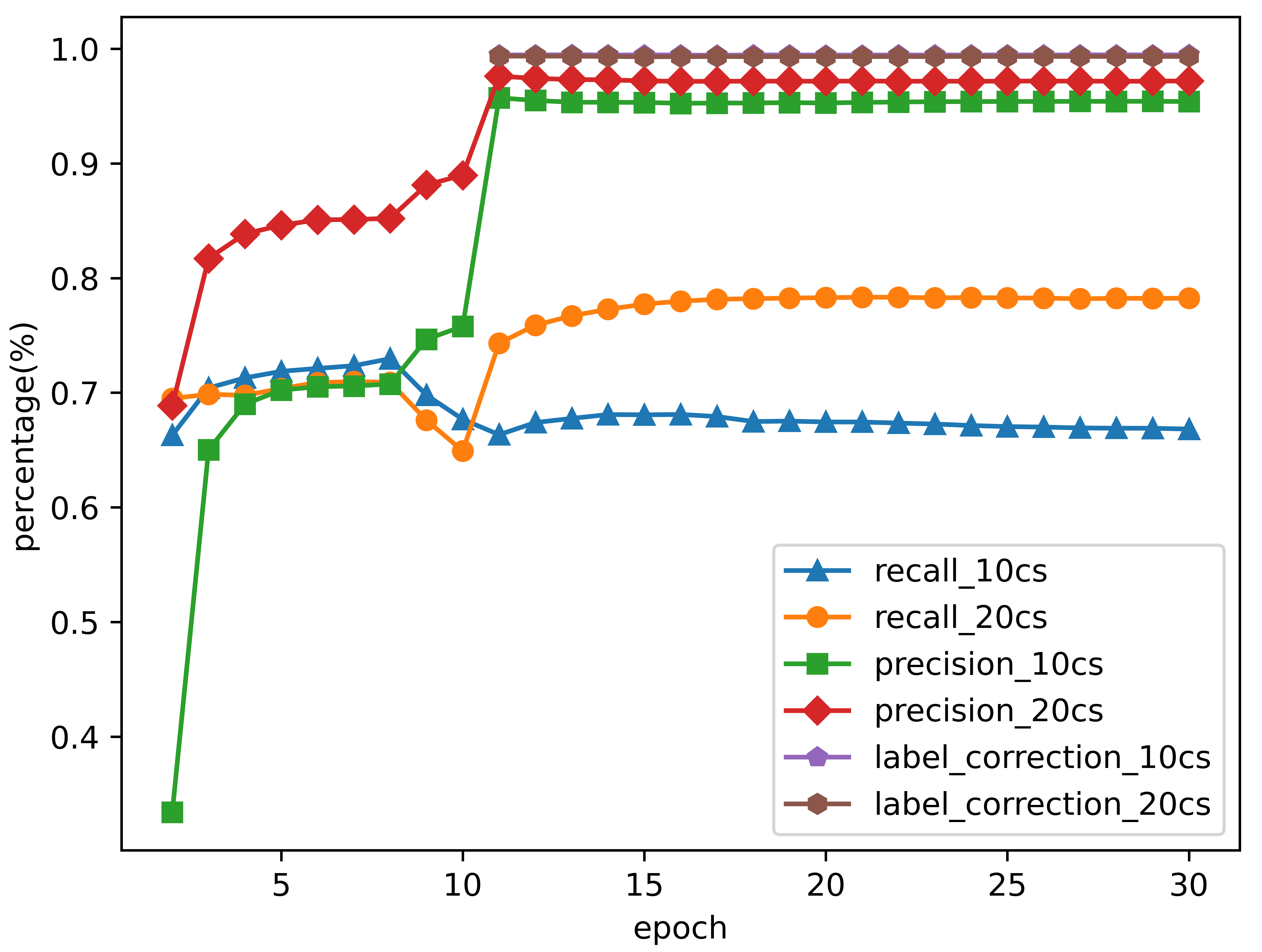}
	\caption{The recall and precision of label noise detection, and the label correction accuracy on 10\% and 20\% closed-set noise datasets.}
	\label{recall_etc}
\end{figure}

During the training phase, the methodology proposed in this study addresses closed-set label noise by continuously detecting and correcting it. Throughout the training iterations on the synthetic closed-set noise dataset, our framework meticulously tracks instances where closed-set noise samples are identified, along with the categories for their correction upon detection. This meticulous record-keeping enables us to quantify the efficacy of our framework in terms of noise detection and the accuracy of label correction. Figure \ref{recall_etc} visually represents the recall, precision, and accuracy of label correction for noisy samples in both the 10\% and 20\% CASIA-WebFace~\cite{yi2014learning} datasets. It is apparent from the figure that our framework exhibits remarkably high precision in detecting noise samples, while achieving label correction accuracy close to 1. This demonstrates the exceptional efficiency of our approach in noise detection and subsequent correction. Upon examination of the recall rate, it is notable that it does not reach particularly high levels. This is attributed to some label noise being identified as ambiguous samples within our framework. Nonetheless, our framework undergoes comprehensive training facilitated by LRF, thereby showcasing robustness against closed-set noise data in general.

\section{Algorithm of our pipeline}
Our overall algorithmic framework is illustrated in Algorithm 1.
\begin{algorithm}[t]
	\small
	\caption{RepFace}\label{algorithm}
	\textbf{Input}: The feature of i-th sample $x_i$ with it's label $y_i$ and index $Idx_i$, last fully-connected layer parameter $W$, embedding network parameter $\Theta$, current epoch $T_k$, start epoch $T_s$.
	\textbf{Output}: $W$, $\Theta$.
	\begin{algorithmic}[1]
		\WHILE{not converged}
		\STATE Generate auxilary samples and calculate noise indicator Eq.\ref{Indicator};
		\IF{$T_k < T_s$}
		\STATE Train model after clean noisy samples;
		\ELSE 
		\IF{$d_i > \tau$}
		\STATE Calculate smoothing parameter $k = Sigmoid(10\cdot d_i)$;
		\STATE calculate Smoothing label Eq.\ref{hardcorrectlb} and Smoothing hardming Eq.\ref{hardcorrect};
		\ENDIF
		\IF{$0<d_i<\tau$}
		\STATE Update record model predict infomation $cos\theta_{y_j}$ Eq.\ref{lfcos}; 
		\STATE Calculate fused labels $q_r$ of ambiguous samples Eq.\ref{lflabel}; 
		\ENDIF
		\STATE Calculate loss $\mathcal{L}_{RepFace}$ Eq.\ref{prog};
		\ENDIF
		\STATE Update the parameter $W$ and $\Theta$ by Stochastic Gradient Descent (SGD) ;
		\ENDWHILE
	\end{algorithmic}
\end{algorithm}

\end{document}